\pdfoutput=1

\documentclass[11pt]{article}
\usepackage{EMNLP2023}
\usepackage{times}
\usepackage{latexsym}
\usepackage{multirow}
\usepackage[normalem]{ulem}
\useunder{\uline}{\ul}{}
\usepackage[T1]{fontenc}

\usepackage[utf8]{inputenc}

\usepackage{microtype}
\usepackage{inconsolata}

\usepackage{graphicx}

\usepackage{multirow}
\usepackage{array}

\usepackage{enumitem}

\title{DISC-LawLLM: Fine-tuning Large Language Models \\ for Intelligent Legal Services}

\author{
    Shengbin Yue\textsuperscript{\rm 1,\rm 2},
    Wei Chen\textsuperscript{\rm 3},
    Siyuan Wang\textsuperscript{\rm 4},
    Bingxuan Li\textsuperscript{\rm 4},
    Chenchen Shen\textsuperscript{\rm 4},
    Shujun Liu\textsuperscript{\rm 4}, \\
    \textbf{Yuxuan Zhou\textsuperscript{\rm 4}, 
    Yao Xiao\textsuperscript{\rm 7},
    Song Yun\textsuperscript{\rm 6},
    Xuanjing Huang\textsuperscript{\rm 5},
    Zhongyu Wei\textsuperscript{\rm 1,\rm 4}\thanks{Corresponding Author}}
    \\[.3cm]
    \textsuperscript{\rm 1}Research Institute of Intelligent Complex Systems, Fudan University, China \\
    \textsuperscript{\rm 2}Shanghai Center For Mathematical Sciences, Fudan University, China \\
    \textsuperscript{\rm 3}{School of Software Engineering, Huazhong University of Science and Technology, China} \\
    \textsuperscript{\rm 4}{School of Data Science, Fudan University, China} \\
    \textsuperscript{\rm 5}{School of Computer Science, Fudan University, China} \\ 
    \textsuperscript{\rm 6}{Rule of Law Institute, Northwest University of Political and Law, China} \\
    \textsuperscript{\rm 7}{New York University Shanghai, China}
    \\
    \{sbyue23,bxli16,ccshen22,yxzhou23\}@m.fudan.edu.cn, 1171991@s.hlju.edu.cn,
    yx2436@nyu.edu, \\
    \{chenwei18,wangsy18,shujunliu20,xjhuang,zywei\}@fudan.edu.cn
}

\begin{document}
\maketitle
\begin{abstract}
We propose DISC-LawLLM, an intelligent legal system utilizing large language models (LLMs) to provide a wide range of legal services. We adopt legal syllogism prompting strategies to construct supervised fine-tuning datasets in the Chinese Judicial domain and fine-tune LLMs with legal reasoning capability. We augment LLMs with a retrieval module to enhance models' ability to access and utilize external legal knowledge. A comprehensive legal benchmark, DISC-Law-Eval, is presented to evaluate
intelligent legal systems from both objective and subjective dimensions. Quantitative and qualitative results on DISC-Law-Eval demonstrate the effectiveness of our system in serving various users across diverse legal scenarios.
The detailed resources are available at \emph{\url{https://github.com/FudanDISC/DISC-LawLLM}}.
\end{abstract}

\section{Introduction}
\label{sec:introduction}

With the rise of legal artificial intelligence (LegalAI)~\cite{gardner1987artificial, zhong2020does}, the legal domain is undergoing significant transformation. Through automating legal tasks including legal information extraction~\cite{bommarito2018lexnlp}, interactive argument pair extraction~\cite{ji2018incorporating,ji2019discrete,yuan2021overview},  case retrieval~\cite{ma2021lecard}, judgment prediction~\cite{song2021inferring,ye-etal-2018-interpretable, yang2019legal}, and legal question answering~\cite{kien2020answering}, intelligent legal systems benefit various groups of people. It boosts the efficiency of legal professionals by reducing the heavy burden of paperwork, and simplifies access to legal services and remote legal advice for general populations. Besides, it offers invaluable assistance to students in their legal knowledge pursuits and examinations.


\begin{figure*}[h]
    \centering
    \includegraphics[width=\linewidth]{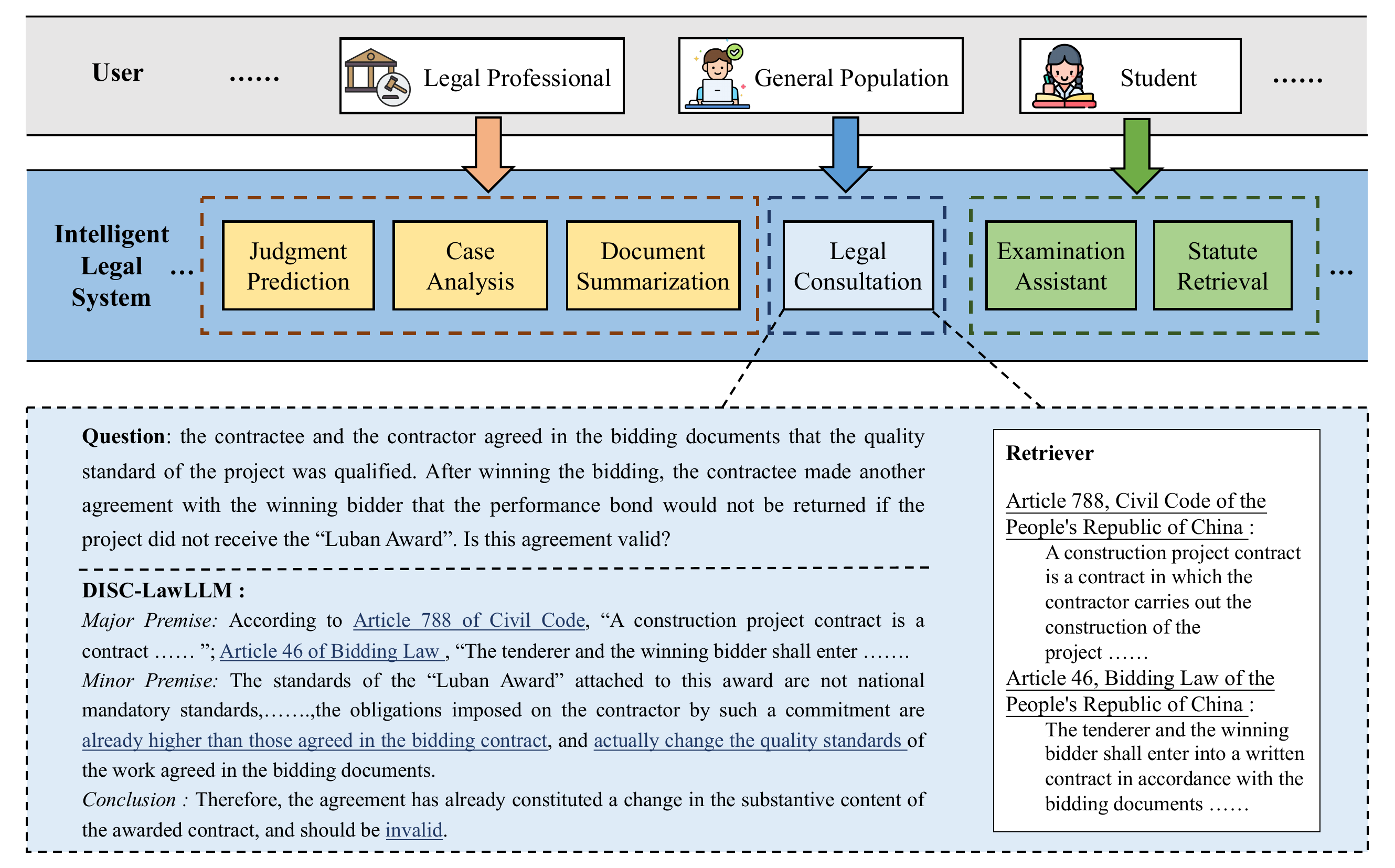}
    \caption{Overview of DISC-LawLLM serving different users in different legal scenarios. For an example of legal consultation, DISC-LawLLM utilizes the retrieved statutes to perform legal reasoning (i.e., the statute is the major premise, the fact in the question is the minor premise, and the final inference is the conclusion).}
    \label{fig:legal-advice-ex}
\end{figure*}

Prior work on LegalAI mainly focuses on specific tasks by constructing datasets and presenting machine learning algorithms. Although they demonstrate some impressive results, this approach confines the broader application of intelligent legal systems. Recent advancements in large language models (LLMs)~\cite{chatgpt,gpt4,BLOOMZ,llama} have shown remarkable instruction-following ability across varied domains, offering tremendous potential for the legal sector. 
Some initial progress has been made \cite{lawyerllama, chatlaw} by fine-tuning general LLMs to utilize legal knowledge for simple question answering, however, legal services are far more complicated and beyond dialogue.

As shown in Figure \ref{fig:legal-advice-ex}, intelligent legal systems have broad application scenarios, serving different groups of users, including professional legal practitioners, everyday individuals seeking legal advice, and law students pursuing academic achievement.
For legal practitioners, the system should provide advanced legal tools for statute retrieval, case analysis, and document summarization.
For the general public, the system should be able to offer legal consultation for statute interpretation and dispute resolution. 
For law students, the system serves as a tutor, helping to consolidate legal knowledge and providing solutions to exam questions.

An example for legal consultation is illustrated in Figure \ref{fig:legal-advice-ex}, where the intelligent legal system can leverage law knowledge to mine facts from the inquiry, and deduce the conclusion to provide legal services. This highlights two primary challenges for building the system. (1) High demand of reasoning ability for legal issues. The legal assistance process is specialized and requires intricate legal reasoning. For example, all legal responses in jurisprudence should follow the structure of legal syllogism~\cite{posner1990problems,jiang2023legal}, which involves a major premise representing the legal proposition, a minor premise symbolizing the factual proposition, and a conclusion representing the judgment. (2) In need of retrieval and inference ability of external legal knowledge. It requires a deep understanding and precise reference of legal knowledge for delivering legal interpretation and alleviating the hallucination. Besides, the repository of legal knowledge is constantly evolving with newly emerged and amended regulations. 

To this end, we present DISC-LawLLM, our legal large language model tailored for building intelligent legal systems with legal reasoning and knowledge retrieval capability. We begin by adopting the legal syllogism prompting strategy to construct supervised fine-tuning datasets in the Chinese Judicial domain, named DISC-Law-SFT. These datasets are then employed to train DISC-LawLLM with legal reasoning on top of a general domain Chinese LLM with 13B parameters~\footnote{In this version, we use Baichuan~\citep{baichuan13b}as the base model. Note that our strategy can be applied to all decoder-only foundation models.}. Besides, we also introduce a retrieval module to source up-to-date and precise legal evidence, enhancing the retrieval augmented DISC-LawLLM's ability to generate more reliable responses. 

\begin{figure*}[h]
    \centering
    \includegraphics[width=\linewidth]{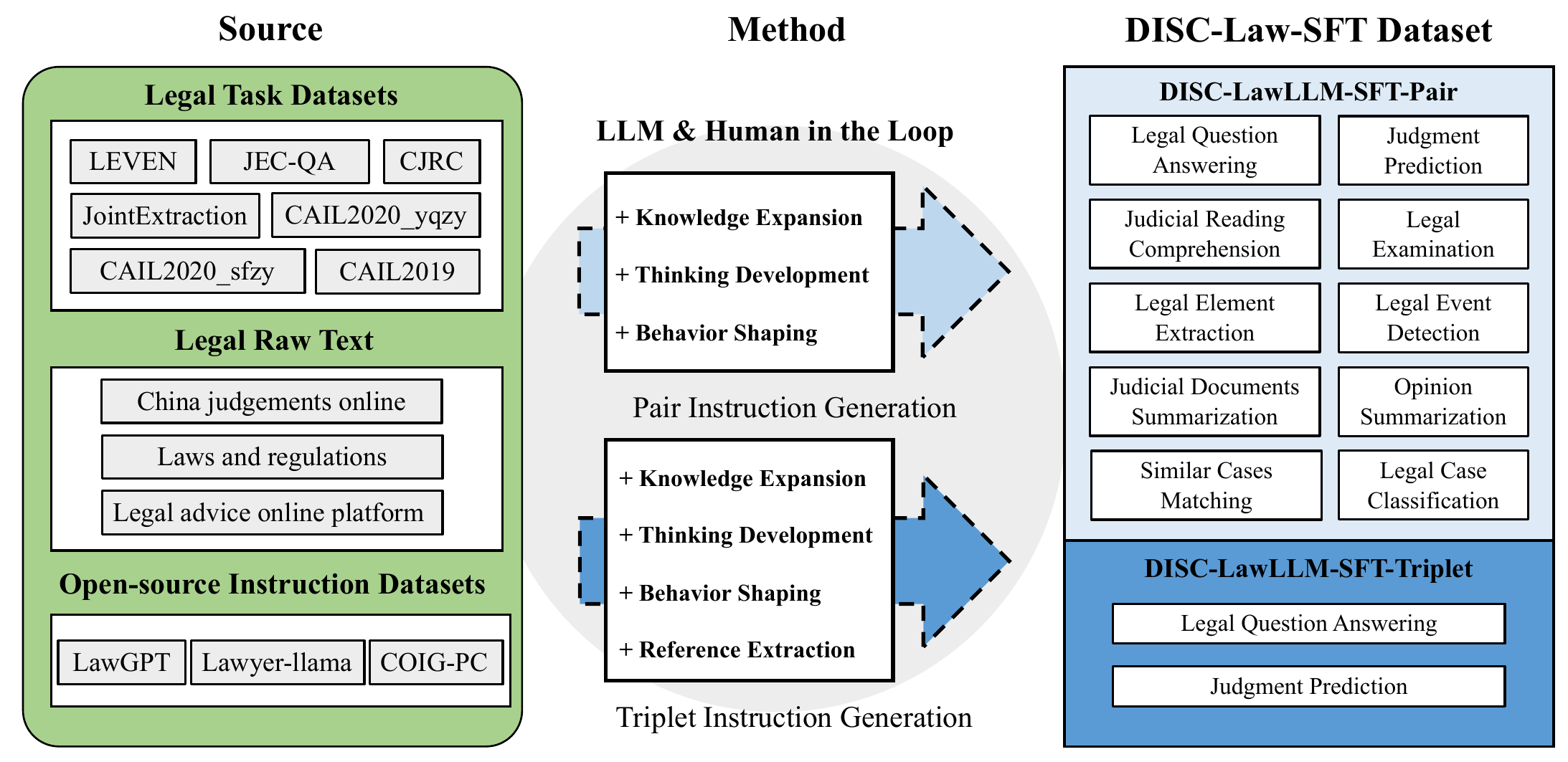}
    \caption{Construction of DISC-Law-SFT Datasets}
    \label{fig:data-process}
\end{figure*}

Finally, we design a legal benchmark, DISC-Law-Eval, to provide a comprehensive assessment of intelligent legal systems from both objective and subjective dimensions. For the objective perspective, our benchmark incorporates multiple-choice with both single and multiple answers sourced from law-related standardized examinations. These questions are categorized into three difficulty levels: \emph{Easy}, \emph{Normal} and \emph{Hard},  
which allows for a deeper insight into the model's grasp of legal knowledge and reasoning capabilities.
For the subjective perspective, we curate a select compilation of high-quality legal Q\&A cases, and utilize GPT-3.5 as an arbitrator to assess the model's metrics in terms of \emph{Accuracy}, \emph{Completeness} and \emph{Clarity}. For each evaluation question, we provide a ground truth to the arbitrator model to reduce potential biases during the assessment phase.

Experimental Results reveal that DISC-LawLLM significantly outperforms existing legal large language models. Even compared to GPT-3.5-turbo \cite{chatgpt} with 165B parameters, DISC-LawLLM excels in a majority of tested subjects of objective evaluation. Our DISC-LawllM equipped with more extensive Chinese legal knowledge and legal reasoning, consistently generates more reliable responses.

\section{Related Work}
\label{sec:related}
\label{sec:datasets}

Large Language Models (LLMs) have achieved astounding performance on different conventional linguistic tasks, demonstrating powerful generality. However, these generic LLMs have proven to be unsuitable for some domain-specific tasks, such as law. This has greatly stimulated researchers' enthusiasm to explore LLMs in the legal domain. Currently, some initial progress has been made in legal LLMs. Specifically, the LaWGPT \cite{lawgpt} series was built on Chinese-LLaMA-7B \cite{chinese-llama-alpaca}, ChatGLM \cite{du2022glm}, and Chinese-alpaca-plus-7B \cite{chinese-llama-alpaca} by training with integrated datasets from the Chinese legal domain and a large-scale Chinese legal corpus enriched with domain-specific terminologies. Lawyer LLaMa \cite{lawyerllama} conducted continuous pre-training on Chinese-LLaMa-13B \cite{chinese-llama-alpaca} and constructed a large number of instruction finetuning datasets to further enhance its ability to provide legal advice. LexiLaw, based on ChatGLM-6B \cite{du2022glm}, was trained with three different methods including LoRA, P-tuning, and finetuning. Additionally, LawGPT\_zh \cite{lawgpt} used self-instruct methods to construct a Q\&A dataset and used LoRA to fine-tune ChatGLM-6B. Chatlaw \cite{chatlaw} was trained based on Ziya-LLaMA-13B-v1 \cite{Fengshenbang-LM} and Anima-33B separately.
Previous work has focused on dialogue competence, one of the intelligent justice tasks. Different from them, we propose an intelligent legal system to provide a wide range of legal services.

\section{DISC-Law-SFT Datasets}
\begin{table*}[ht]
\centering
\setlength{\tabcolsep}{1.5mm}
\resizebox{0.94\textwidth}{!}{
{\begin{tabular}{cccc}
\hline
Dataset                           &   Task                   &  \ \ \ Size \ \ \  & Scenario
\\ \hline
\multirow{10}{*}{DISC-Law-SFT-Pair}        
                                  & Legal Element Extraction         & 32K  & \multirow{7}{*}{Legal Professional Tools} \\
                                  & Legal Event Detection            & 27K  \\
                                  & Legal Case Classification        & 20K  \\
                                  & Judgement Prediction             & 11K  \\
                                  & Similar Cases Matching           & 8K   \\
                                  & Documents Summarization & 9K   \\
                                  & Public Opinion Summarization     & 6K   \\ 
                                  & Legal Question Answering         & 93K  & Legal Consultation \\
                                  & Document Reading Comprehension   & 38K & \multirow{2}{*}{Examination Assistant}\\
                                  & Judicial Examination             & 12K  \\
                                  \hline
\multirow{2}{*}{DISC-Law-SFT-Triplet} & Judgement Prediction             & 16K  & Legal Professional Tools \\ 
                                  & Legal Question Answering         & 23K  & Legal Consultation \\
                                  \hline
\multirow{2}{*}{General}          & Alpaca-GPT4                      & 48K    & \multirow{2}{*}{General} \\
                                  & Firefly                          & 60K  \\ \hline
Total                             &                                  & 403K \\ \hline
\end{tabular}}
}
\caption{Statistics of DISC-Law-SFT Dataset.}
\label{tab:SFT-size}
\end{table*}
To train DISC-LawLLM, we construct a high-quality supervised fine-tuning dataset, DISC-Law-SFT with two subsets, namely DISC-Law-SFT-Pair and DISC-Law-SFT-Triplet. The former part aims to introduce the legal reasoning ability to the LLM, while the later part help to improve the model's ability of utilizing external knowledge. The workflow of constructing DISC-Law-SFT is shown in Figure.~\ref{fig:data-process}.

Legal intelligent applications in different scenarios usually require combinations of multiple fundamental capabilities of legal text understanding and generating. To this end, we construct instruction samples converging a range of legal tasks, including legal information extraction, judgment prediction, document summarization, and legal question answering, ensuring coverage of diverse scenarios. General LLMs and human labeller are involved to re-construct original samples to generate instructions in two forms of the pair (<input, output>) and the triplet (<input, output, reference>).

\subsection{Data Sources}

We obtain original samples from three sources, namely, public NLP legal task datasets, legal raw text and open-source instruction datasets. 

\paragraph{1) Public NLP Legal Task Datasets.}
Public datasets covers a range of legal NLP tasks and provide human annotations which can be utilized to generate high quality instructions. We collect public datasets of specific legal tasks related to Chinese justice, including Legal Information Extraction (LEVEN \cite{yao-etal-2022-leven} and JointExtraction \cite{chen-etal-2020-joint-entity}), Legal Text Summarization (CAIL2020-sfzy~\cite{CAIL2020} and CAIL2022-yqzy~\cite{CAIL2022}), Legal Question Answering (JEC-QA \cite{JEC-QA} and CJRC \cite{CJRC}), and Judgement Prediction (CAIL2018 \cite{CAIL2018-paper}). 

\paragraph{2) Legal Raw Text.}
In order to include more scenarios of legal services, we explore to generate instructions from legal raw text. We crawl up an expansive collection of real-world legal text to construct instruction data. This includes consultation data from judicial advisory websites, Chinese laws and regulations, typical cases, judicial verdicts and law-related examinations.

\paragraph{3) Open-source Instruction Datasets.}
In addition, we also borrow some samples from recently opened instruction datasets. We collect open-source instruction data, including Lawyer-LLaMa \cite{lawyerllama}, LawGPT-zh \cite{LAWGPT-zh} and COIG-PC \cite{zhang2023chinese}. 

\subsection{Pair Instruction Generation}
\label{pair_generation}

To construct instructions for supervised fine-tuning DISC-LawLLM, we first use rule-based methods to clean the data and transform it into ``input-output'' pairs. However, these pairs are too rigid and noisy in linguistic patterns and the expression styles can differ across sources. Therefore, we reconstruct the instruction pairs using the following three methods with the assistance of general large language model.

\begin{figure*}[h]
    \centering
    \includegraphics[width=\linewidth]{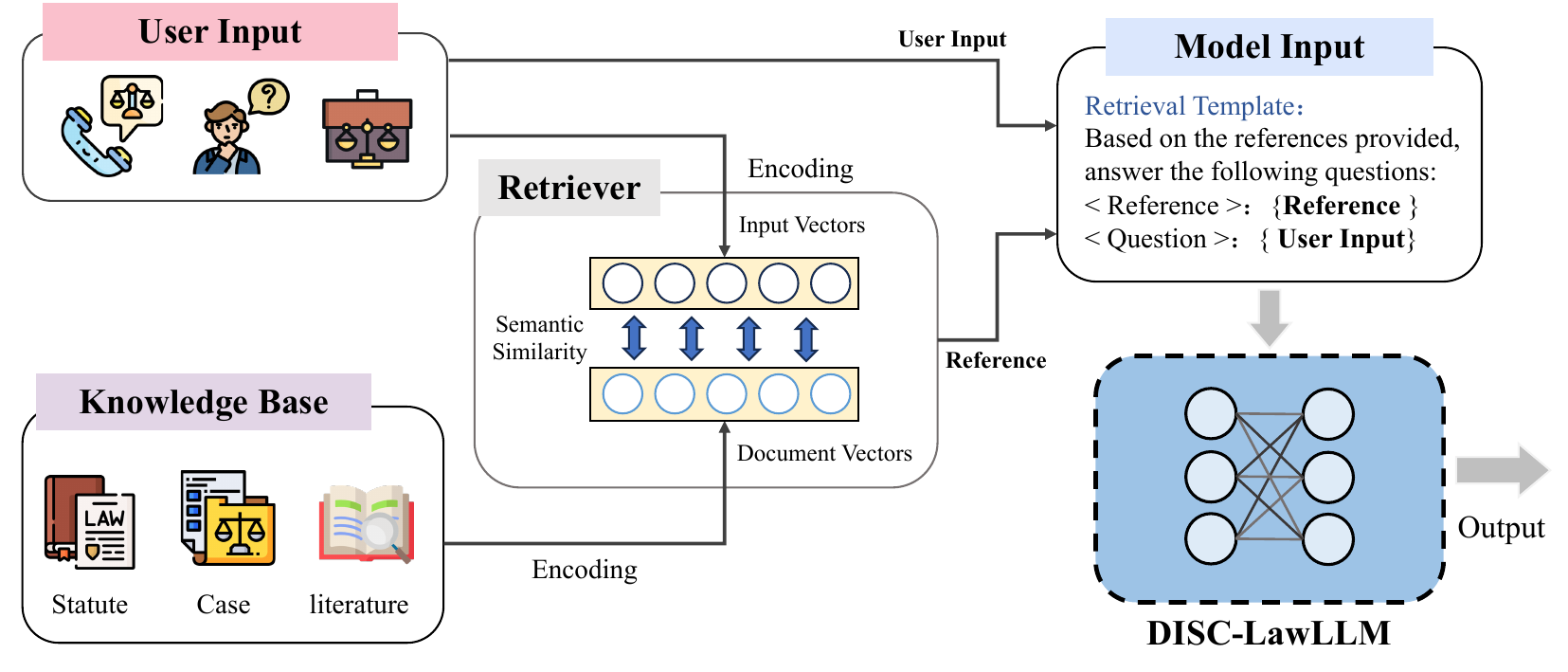}
    \caption{Overview of Retrieval Augmented DISC-LawLLM. Specifically, the reference related to user input in the knowledge base is first retrieved, and then the reference and user input are fed into the DISC-LawLLM with retrieval behavior.}
    \label{fig:ref}
\end{figure*}
\paragraph{Behavior Shaping.}
In the syllogism of legal judgment, the major premise is the applicable law, while the minor premise is pertinent facts, and the conclusion is the final judgment. This constitutes a foundational legal reasoning process for judges. Every case can culminate in a conclusion articulated through a syllogism, as outlined below:
\begin{itemize}[nosep]
    \item Major premise: laws
    \item Minor premise: pertinent facts 
    \item Conclusion: judgment
\end{itemize}
Inspired by legal syllogism prompting \cite{jiang2023legal} and self-construct \cite{wang2022self}, we utilize LLMs to refine output responses for consistency with legal syllogism. We design prompts for GPT-3.5-turbo, to ensure that each conclusion should be drawn from laws and pertinent facts, and responses should be in Chinese. 
\label{sec:Reference}
\label{sec:datasets}

\paragraph{Knowledge Expansion.}
For multiple-choice questions where behavior shaping is not applicable for selecting an option, we directly expand output responses with legal knowledge to provide more reasoning details. These questions come from various Chinese law-related exams and knowledge competitions, involving knowledge of criminal law, constitutional law, and civil law. While many of them only offer answer options, we use LLMs to expand the involved legal knowledge given the correct answer and reconstruct instruction pairs.

\paragraph{Thinking Development.}
Chain of Thought (CoT) has been proven effective in enhancing the reasoning ability of models. To further endow legal reasoning into the model, we devise law-specific chains of thought, termed LCoT, to enforce the model conduct legal syllogism to derive the answer. LCoT incorporates prompts that transform input $X$ into $X_{l}$ as follows: 

\vspace{2mm}
\emph{In the legal syllogism, the major premise is articles of law, the minor premise is the facts of the case, and the conclusion is the judgment of case.}

\emph{Case: $X$}

\emph{Let us use legal syllogism to think and output the judgment:}
\vspace{2mm}



\subsection{Triplet Instruction Generation}
To generate supervised instruction triplets <input, output, reference> for retrieval augmented DISC-LawLLM, we create a subset called DISC-Law-SFT-Triplet. 
For each entry, we utilize the three strategies outlined in Sec.~\ref{pair_generation} to process the original data and obtain the input and output. Subsequently, we design heuristic rules to extract reference information from this raw data. 

\subsection{Dataset Overview} 
Our DISC-Law-SFT dataset consists of more than 10 tasks, such as Legal Element Extraction, Case Matching, Judgment Prediction, Document Summarization, and Question Answering, covering a diverse range of legal scenarios.
Additionally, we incorporate general instruction data to enrich the diversity of our training set, mitigating the risk that foundational capability diminishes during the SFT training phase in the legal domain. 
Specifically, we sourced over 100k samples from alpaca\_gpt4\_data\_zh \cite{peng2023instruction} and Firefly \cite{Firefly}.  
Detailed statistics of our datasets are provided in Table \ref{tab:SFT-size}.

\section{DISC-LawLLM}
\label{sec:training}

To build an intelligent legal system with legal reasoning and retrieval ability, we form our DISC-LawLLM using two steps, Supervised Fine-Tuning (SFT) and Retrieval Augmentation.
\begin{figure*}[h]
    \centering
    \includegraphics[width=\linewidth]{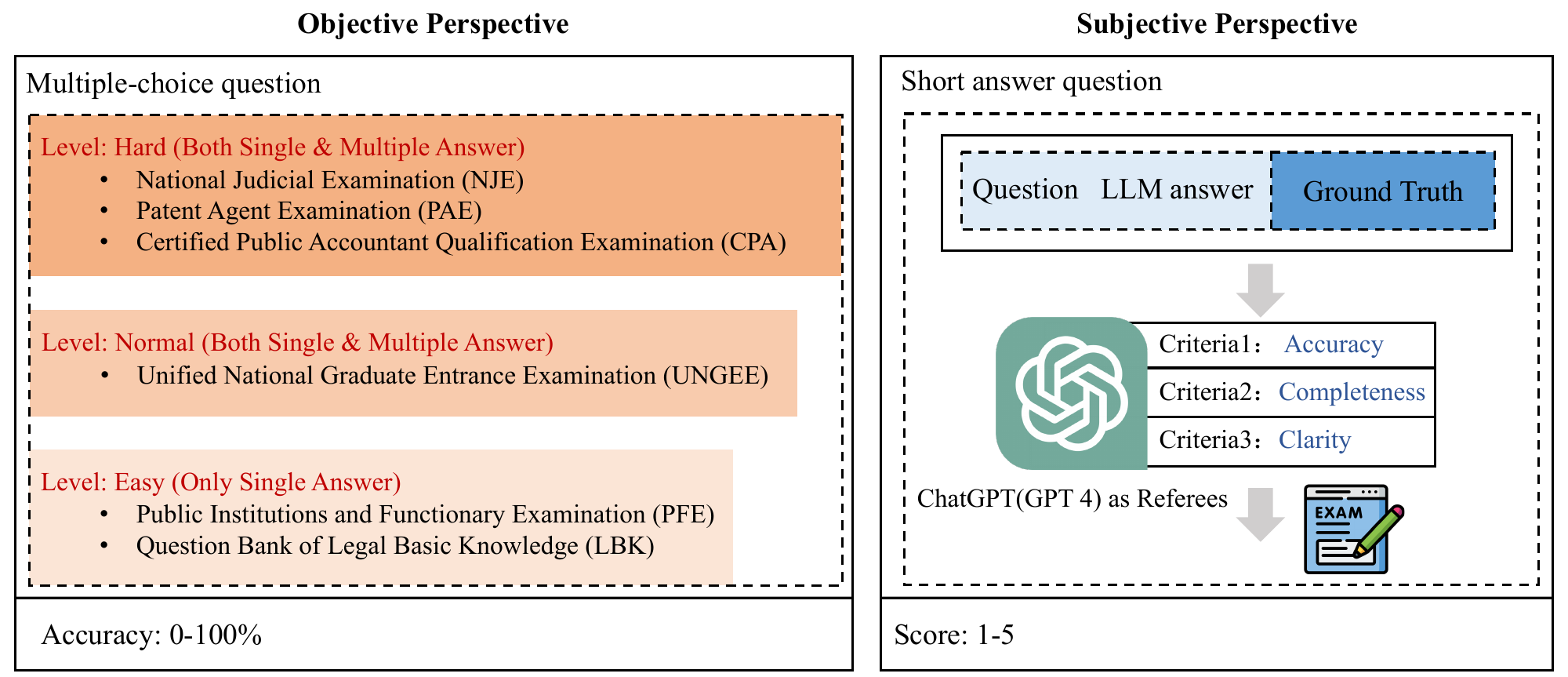}
    \caption{Overview of DISC-Law-Eval Benchmark, assessing systems from both objective and subjective perspectives.}
    \label{fig:benchmark}
\end{figure*}

\subsection{Supervised Fine-Tuning}
We first develop our DISC-LawLLM on top of the Baichuan-13B-Base model \cite{baichuan13b}, which is an open-source LLM with over 13.2 billion parameters that was trained on 1.4 trillion tokens corpus, exhibiting ideal performance in both English and Chinese.
Specifically, we perform supervised fine-tuning using our DISC-Law-SFT dataset. This refined SFT data enabled the model to be equipped with legal reasoning and judicial behavioral patterns.

The hyperparameters setting of this training process are as follows: global batch size of 256, learning rate of 5e-5, 2 epochs training stage, maximum source length of 2048 tokens, maximum target length of 1024 tokens. The training process was carried out on 8*A800 GPUs and the training cost is further reduced with the help of deepspeed \cite{rasley2020deepspeed}. 

\subsection{Retrieval Augmentation}
In many legal scenarios, such as legal consultation and judgment prediction, users expect model's responses to be strongly supported by legal precedents and statutes. While we fine-tune LLM with high-quality instruction data, it might produce inaccurate responses due to hallucinations or outdated knowledge. To address this, we augment the DISC-LawLLM with a retrieval module based on an open-source retrieval framework~\footnote{\url{https://github.com/chatchat-space/Langchain-Chatchat}}. 

We begin by building a knowledge base with over 50 categories of Chinese laws, including the Constitution, Criminal Law, Administrative Procedure Law, Copyright Law, Patent Law. We encode these laws as vectors and save them locally. 
Given a user input, our retriever then returns Top-K most relevant documents from the knowledge base by calculating their similarity to the input. These candidate documents, along with the user input, are formulated using our designed template and then fed into the DISC-LawLLM. By querying the knowledge base for references, the model can better understand the major premise, leading to more accurate and reliable answers.  

To adapt to retrieval scenarios, we specifically employ DISC-Law-SFT-Triplet, as mentioned in Section 3.3, as our SFT dataset for training. This enables the model to infer reliable results using retrieved references.
In addition, our knowledge base is designed for dynamic updates, ensuring the availability of up-to-date laws. Therefore, our thinking-developed DISC-LawLLM can deduce the correct answer based on the new knowledge retrieved.

\label{sec:Reference}
\label{sec:datasets}
\section{DISC-Law-Eval Benchmark}

There is no established benchmark to provide a comprehensive assessment of intelligent legal systems. Inspired by the composition of the bar exam, as shown in Figure \ref{fig:benchmark}, we develop a fair evaluation framework, DISC-Law-Eval Benchmark, assessing systems from both the objective perspective and subjective perspective.
\begin{table*}[]
\centering
\setlength{\tabcolsep}{0.75mm}
{\begin{tabular}{c|c|cccccc|cc|cc|c}
\hline
\multirow{3}{*}{Model} & \multirow{3}{*}{Size} & \multicolumn{6}{c|}{Hard}                                                                                                                     & \multicolumn{2}{c|}{Normal}     & \multicolumn{2}{c|}{Easy}                            & \multirow{3}{*}{Average} \\ \cline{3-12}
                       &                       & \multicolumn{2}{c|}{NJE}                             & \multicolumn{2}{c|}{PAE}                             & \multicolumn{2}{c|}{CPA}        & \multicolumn{2}{c|}{UNGEE}      & \multicolumn{1}{c|}{PFE}            & LBK            &                          \\ \cline{3-12}
                       &                       & S              & \multicolumn{1}{c|}{M}              & S              & \multicolumn{1}{c|}{M}              & S              & M              & S              & M              & \multicolumn{1}{c|}{S}              & S              &                          \\ \hline
ChatGLM                & 6B                    & 31.66          & \multicolumn{1}{c|}{1.08}           & 27.97          & \multicolumn{1}{c|}{2.90}           & 37.06          & 13.33          & 39.69          & 20.69          & \multicolumn{1}{c|}{37.65}          & 42.91          & 24.66                    \\
Baichuan-Chat          & 13B                   & 31.47          & \multicolumn{1}{c|}{10.15}          & 29.66          & \multicolumn{1}{c|}{8.70}           & 35.53          & 19.17          & 50.0           & {\ul 27.59}          & \multicolumn{1}{c|}{53.12}          & 53.45          & 30.78                    \\
Chinese-alpaca2        & 13B                   & 25.7           & \multicolumn{1}{c|}{10.15}          & 30.51          & \multicolumn{1}{c|}{11.59}          & 32.99          & 19.17          & 40.94          & 21.84          & \multicolumn{1}{c|}{44.12}          & 43.27          & 26.73                    \\
GPT-3.5-turbo          & 175B                  & {\ul 36.5}     & \multicolumn{1}{c|}{{\ul 10.58}}          & {\ul 37.29}    & \multicolumn{1}{c|}{{\ul 17.03}}    & \textbf{42.13} & \textbf{21.67} & \textbf{51.25} & \textbf{28.74} & \multicolumn{1}{c|}{{\ul 53.53}}    & {\ul 54.18}    & {\ul 34.10}                    \\ \hline
LexiLaw                & 6B                    & 20.11          & \multicolumn{1}{c|}{7.56}           & 23.73          & \multicolumn{1}{c|}{10.14}          & 24.87          & 19.17    & 31.56          & 16.09          & \multicolumn{1}{c|}{31.76}          & 40.36          & 21.50                     \\
LawGPT                 & 7B                    & 22.91          & \multicolumn{1}{c|}{6.26}           & 31.36          & \multicolumn{1}{c|}{7.61}           & 25.38          & 16.67          & 30.31          & 13.79          & \multicolumn{1}{c|}{34.71}          & 29.09          & 20.60                    \\
Lawyer LLaMA           & 13B                   & 35.75          & \multicolumn{1}{c|}{5.62}           & 32.20          & \multicolumn{1}{c|}{6.52}           & 29.95          & 13.33          & 32.50          & 14.94          & \multicolumn{1}{c|}{39.41}          & 39.64          & 25.05                    \\
ChatLaw                & 13B                   & 27.56          & \multicolumn{1}{c|}{7.99}           & 31.36          & \multicolumn{1}{c|}{9.42}           & 35.53          & 11.67          & 35.62          & 17.24          & \multicolumn{1}{c|}{42.35}          & 41.09          & 25.20                    \\ \hline
DISC-LawLLM            & 13B                   & \textbf{42.09} & \multicolumn{1}{c|}{\textbf{19.87}} & \textbf{40.68} & \multicolumn{1}{c|}{\textbf{18.48}} & {\ul 39.59}    & {\ul 19.17}    & {\ul 50.94}    & 25.29          & \multicolumn{1}{c|}{\textbf{57.06}} & \textbf{54.91} & \textbf{37.10}                    \\ \hline
\end{tabular}
}
\caption{Results compared with general and legal LLMs on Objective Evaluation. Bold represents the best result and underlining represents the second best result. S and M are shorthand of single-answer and multiple answers, respectively.}
\label{tab:result-obj}
\end{table*}
\subsection{Objective Evaluation}
\begin{table}[]
\centering
\setlength{\tabcolsep}{2.5mm}
{
\begin{tabular}{c|c|ccc}
\hline
Subject & Level & S & M & Total \\ \hline
CPA & \multirow{3}{*}{Hard} & 197 & 120 & 317 \\
NJE &  & 537 & 463 & 1000 \\
PAE &  & 118 & 276 & 394 \\ \hline
UNGEE & Normal & 320 & 87 & 407 \\ \hline
LBK & \multirow{2}{*}{Easy} & 275 & - & 275 \\
PFE &  & 170 & - & 170 \\ \hline
\end{tabular}
}
\caption{Details of Objective Question Dataset, where S and M are shorthand of single-answer and multiple answers, respectively.}
\label{tab:eval-obj}
\end{table}
To objectively and quantitatively assess the legal knowledge and reasoning capabilities of intelligent legal systems, we design an objective evaluation dataset.
It consists of multiple-choice questions, and each may have one or multiple correct answers.
It can provide a more challenging and reliable measure of whether the model can use its knowledge to reason toward correct answers. We calculate the accuracy to indicate the performance. 


We collect multi-choice questions from a range of Chinese legal standardized examinations and knowledge contests, including National Judicial Examination (NJE), Patent Agent Examination (PAE), Certified Public Accountant Qualification Examination (CPA), Unified National Graduate Entrance Examination (UNGEE), Public Institutions and Functionary Examination (PFE) and Question Bank of Legal Basic Knowledge (LBK). 
According to content complexity and deduction difficulty, we categorize these questions into three levels: \textit{Hard}, \textit{Normal} and \textit{Easy}. 
Considering that many legal LLMs use JEC-QA\cite{JEC-QA} (2007-2017 National Judicial Examination) for their training datasets, our NJE contains a manual collection of test questions during 2018-2022, ensuring a fair evaluation. Table \ref{tab:eval-obj} shows the details of the objective question dataset.

We conduct the objective evaluation in a few-shot setting (4-shot for single-answer questions and 5-shot for multi-answer questions). We use a regular matching method to extract answers from the LLM output, and subsequently compare them to the ground truth to calculate accuracy.

\subsection{Subjective Evaluation}

We further conduct a subjective evaluation to explicitly demonstrate the model's command over legal knowledge and reasoning ability. 
We adopt a question-answering paradigm for this assessment, simulating the process of subjective examination questions. 
We manually construct a high-quality test set from legal consultations, online postings, justice-related publications, and legal documents, comprising 300 examples.
These examples cover scenarios including legal tools, legal consultations, and judgment prediction.

\begin{table*}[]
\centering
\setlength{\tabcolsep}{4mm}
{
\begin{tabular}{cc|ccc|c}
\hline
Model                                & Size & ACC  & CPL  & CLR  & Average \\ \hline
\multicolumn{1}{c|}{ChatGLM}         & 6B   & 2.64 & 2.75 & 3.23 & 2.87    \\
\multicolumn{1}{c|}{Baichuan-Chat}   & 13B  & 3.22 &\textbf{ 3.34} & 3.18 & 3.25    \\
\multicolumn{1}{c|}{Chinese-Alpaca2} & 13B  & 3.13 & 3.23 & 3.17 & 3.17    \\ \hline
\multicolumn{1}{c|}{LexiLaw}         & 6B   & 3.06 & 2.62 & 3.00 & 2.90    \\
\multicolumn{1}{c|}{LaWGPT}          & 7B   & 3.02 & 2.58 & 2.96 & 2.86    \\
\multicolumn{1}{c|}{Lawyer-LLaMa}    & 13B  & 3.13 & 2.83 & 3.35 & 3.10    \\
\multicolumn{1}{c|}{ChatLaw}         & 13B  & 3.31 & 2.90 & 3.35 & 3.19    \\ \hline
\multicolumn{1}{c|}{DISC-LawLLM}     & 13B  & \textbf{3.46} & 3.12 & \textbf{3.59} & \textbf{3.39}    \\ \hline
\end{tabular}}
\caption{Results compared with general and legal LLMs on Subjective Evaluation, where ACC, CPL and CLR are the shorthand of Accuracy, Completeness and Clarity, respectively.}
\label{tab:result-sub}
\end{table*}

To evaluate this subjective response, we evaluate the model's output by eliciting a referee model. Strong LLM judges like GPT-3.5, GPT-4 align well with controlled and crowdsourced human preferences~\citep{zheng2023judging}. In our evaluation, GPT-3.5 serves as a referee and performs the evaluation by providing a rating score from 1 to 5 for each of the following three criteria: 
\textit{accuracy}, \textit{completeness} and \textit{clarity}.

\begin{itemize}
\item Accuracy: The content and semantics of the pending scored answer should be consistent with reference answer.
\item Completeness: Compared to the reference answer, the pending scored answer does not miss any details in the reference answer. Do not let the length of the pending scored answer influence your judgment.
\item Clarity: Compared to the reference answer, the juridical logic analysis of the pending scored answer is rigorous and clear, and the sentences are well-organized.
\end{itemize}
To reduce the self-bias of the referee model, we provide the ground truth to the referee as well, enabling them to score according to the ground truth. We repeat the scoring for each question and finally get the average score on different dimensions. 

\section{Experiments}
\label{sec:experiments}

To demonstrate the excellence of our model, we compare DISC-LawLLM (without retrieval augmentation)  with instruction-aligned general LLMs and exiting legal LLMs on the DISC-Law-Eval benchmark.
The instruction-aligned general LLMs includes: 1) GPT-3.5-Turbo \cite{chatgpt}; 2) Chatglm-6B \cite{du2022glm}; 3) Baichuan-13B-Chat \cite{baichuan13b};5) Chinese-Alpaca2-13B \cite{chinese-alpaca2}.
The legal LLMs include:
1) LaWGPT \cite{lawgpt}; 2) Lawyer-LLaMa \cite{lawyerllama}; 3) ChatLaw \cite{chatlaw}; 4) LexiLaw \cite{lexilaw}. 

\subsection{Results in Objective Evaluation}

Table \ref{tab:result-obj} shows the objective evaluation performance. 
We can see that DISC-LawLLM surpasses nearly all competing LLMs across all subjects with different difficulty levels. 
Even compared to GPT-3.5-Turbo \cite{chatgpt} with 175B parameters, DISC-LawLLM consistently demonstrates superior performance on most subjects, improving accuracy by an average of 7\%. This illustrates the effectiveness of our DISC-LawLLM in inferring correct answers from questions across a broad range of legal subjects.

Specifically, for NJE, PAE and CPA with higher levels of difficulty, DISC-LawLLM surpasses all LLMs by a large margin on NJE and PAE. 
Especially for multi-answer questions that require more discerning judgment and reasoning, our model achieves an improvement of over 50\% compared to the top-performing GPT-3.5-Turbo on NJE and PAE. 
This indicates that DISC-LawLLM has strong jurisprudential reasoning capabilities. Furthermore, the reason that generic LLMs occasionally outperform legal LLMs is likely to be the lack of few-shot instruction following ability during training.

\subsection{Results in Subjective Evaluation}
In the subjective evaluation, we utilize ChatGPT's comprehension to evaluate the model's performance on short answer questions against the Ground Truth.
From Table \ref{tab:result-sub}, we can see that DISC-LawLLM achieves the best performance on most metrics. 
Compared to Chatlaw \cite{chatlaw}, DISC-LawLLM shows a 6\% increase in average performance. We can conclude that: 1) Leveraging the high-quality DISC-Law-SFT dataset enables DISC-LawLLM to generate more reliable responses, leading to outstanding scores in both ACC and CPL. 2) Through the deliberate cultivation of the model's juridical thinking, the responses from DISC-LawLLM exhibit superior jurisprudential logic.


\section{Applications}
\label{sec:Reference}

Our intelligent legal system, DISC-LawLLM, can serve various users across diverse scenarios.
In this section, we showcase its application examples in three scenarios: legal professional tools, legal consultation, and examination assistant. The corresponding figures are displayed in appendix \ref{sec:appendix}. 

\paragraph{Legal Professional Tools.}
 Our DISC-LawLLM simplifies the work of legal professionals by offering advanced tools for extracting legal elements, detecting legal events, analyzing legal cases, matching similar cases, generating judicial summaries, etc.
 Figure \ref{fig:case2} shows two cases of DISC-LawLLM in legal event detection and legal summarization. In the first case, we see that DISC-LawLLM can extract the event trigger words and corresponding event types. In the second case, DISC-LawLLM can generate an accurate summary of the judicial case. These tools not only streamline judicial event monitoring and accelerate the decision-making process, but also facilitate other intelligent legal tasks. 
 
\paragraph{Legal Consultation.}
Our DISC-LawLLM can offer legal consultation for dispute resolution, which greatly facilitates access to legal services and remote legal counseling for the general public.
Figure \ref{fig:case3} shows two cases of DISC-LawLLM for legal consultations about claims and debts and agreement drafting. In the first case, DISC-LawLLM effectively leverages facts in relevant legal base to provide a reliable response about debt apportionment. In the second example, DISC-LawLLM can offer precise drafting suggestions. These instances demonstrate DISC-LawLLM's sound legal knowledge and reasoning proficiency.

\paragraph{Examination Assistant.}
For law students, our DISC-LawLLM serves as a tutor, helping to consolidate legal knowledge and providing solutions to exam questions. Figure \ref{fig:case1} shows two cases of DISC-LawLLM in legal examinations and legal knowledge solutions. For the first case, our DISC-LawLLM can first predict the correct answer, and simultaneously unpack the rationale for answering this question. For the second case, our DISC-LawLLM can provide detailed legal explanations.
Such ability is very valuable for law students, as DISC-LawLLM can help them gain deeper insights and dissect the answers to previously challenging questions, thereby improving their command of the subject.

In addition, we also show two examples with retrieval results in Figure \ref{fig:case4}, which can be extended for more application scenarios.  
Overall, our DISC-LawLLM can bridge the gap between LLMs and various judicial scenarios, which satisfies the demands of broad populations and has significant application value.
\section{Conclusion}
\label{sec:conclusion}

In this paper, we introduce DISC-LawLLM, an intelligent legal system for offering various legal services. 
Based on public NLP legal task datasets, legal raw text and open-source instruction datasets, we utilize ChatGPT to reconstruct legal responses following legal syllogism for supervised fine-tuning. To enhance the reliability of output responses, we incorporate an external retrieval module into our system. 
Through learning legal reasoning and knowledge retrieval capability, DISC-LawLLM outperforms existing legal LLMs on our constructed legal benchmark. This benchmark provides a comprehensive assessment from both subjective and objective perspectives. 
We release the constructed datasets and model weights to foster further research.

\bibliography{anthology,ref}
\bibliographystyle{acl_natbib}

\newpage
\onecolumn
\appendix
\section{Appendix: Supplementary Figures}
\label{sec:appendix}
\begin{figure*}[!ht]
    \centering
\includegraphics[width=\linewidth]{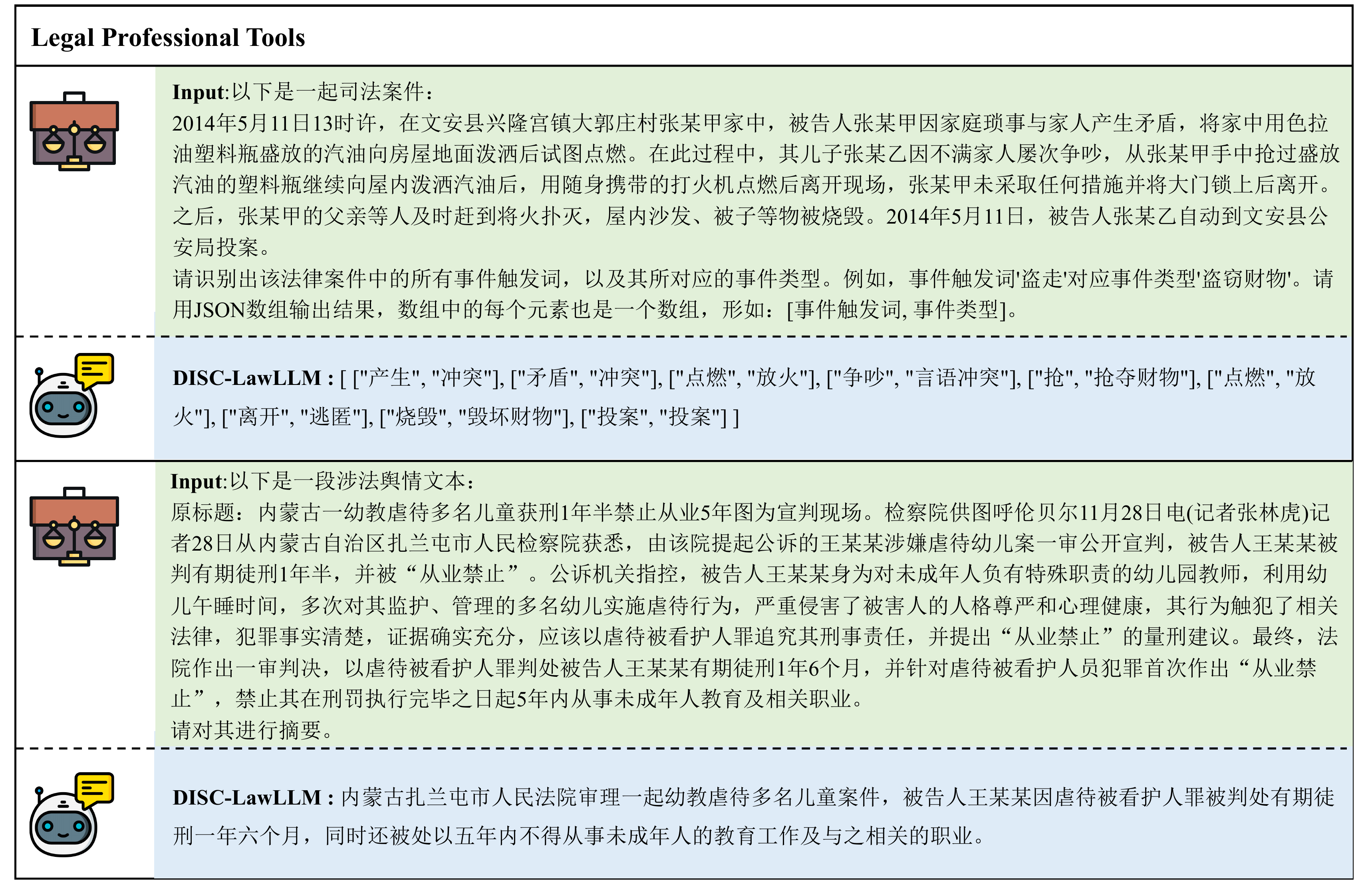}
    \caption{Case of our DISC-LawLLM as legal professional tools}
    \label{fig:case2}
\end{figure*}

\begin{figure*}[!ht]
    \centering
\includegraphics[width=\linewidth]{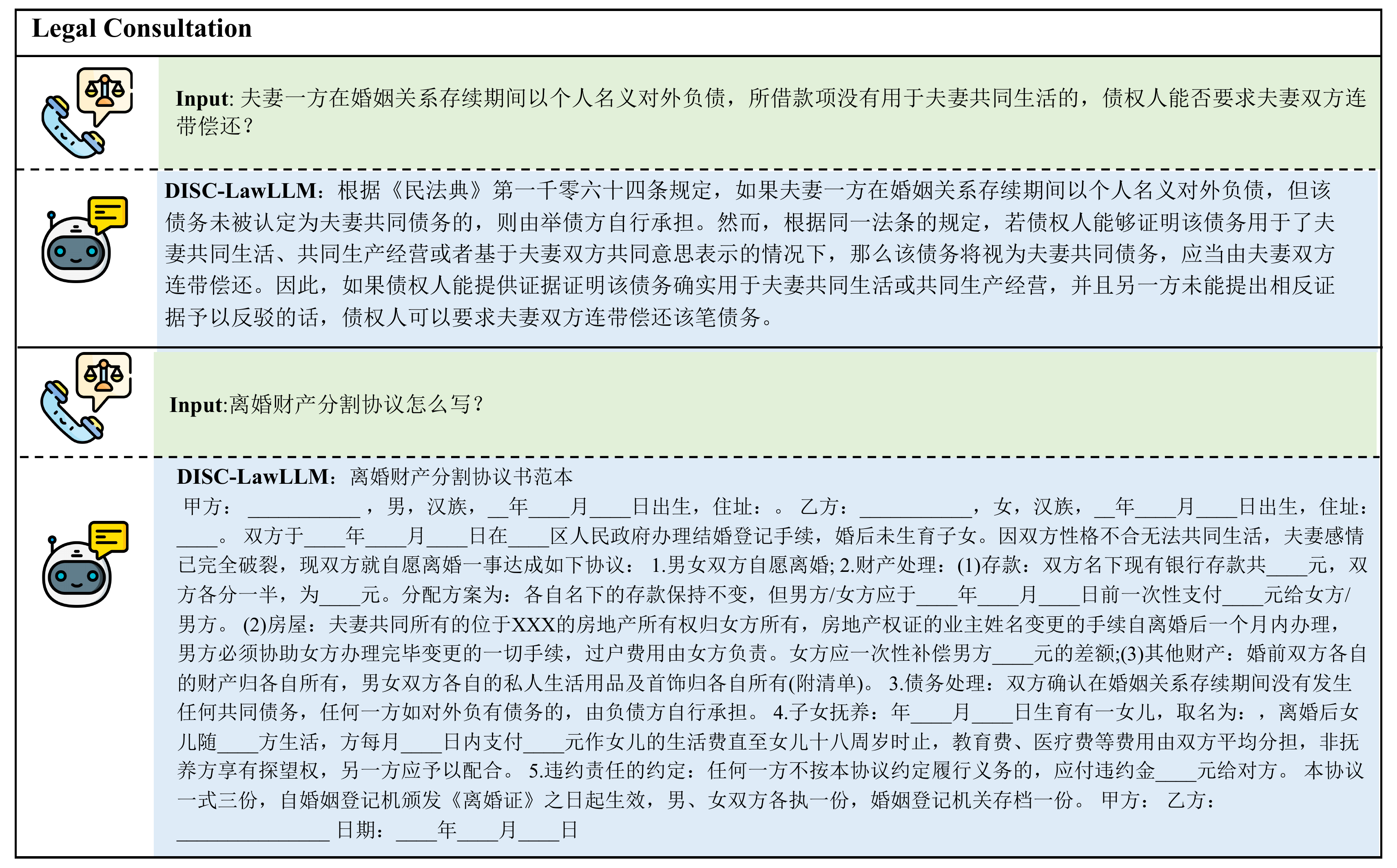}
    \caption{Case of our DISC-LawLLM in legal consultation}
    \label{fig:case3}
\end{figure*}

\begin{figure*}[!ht]
    \centering
\includegraphics[width=\linewidth]{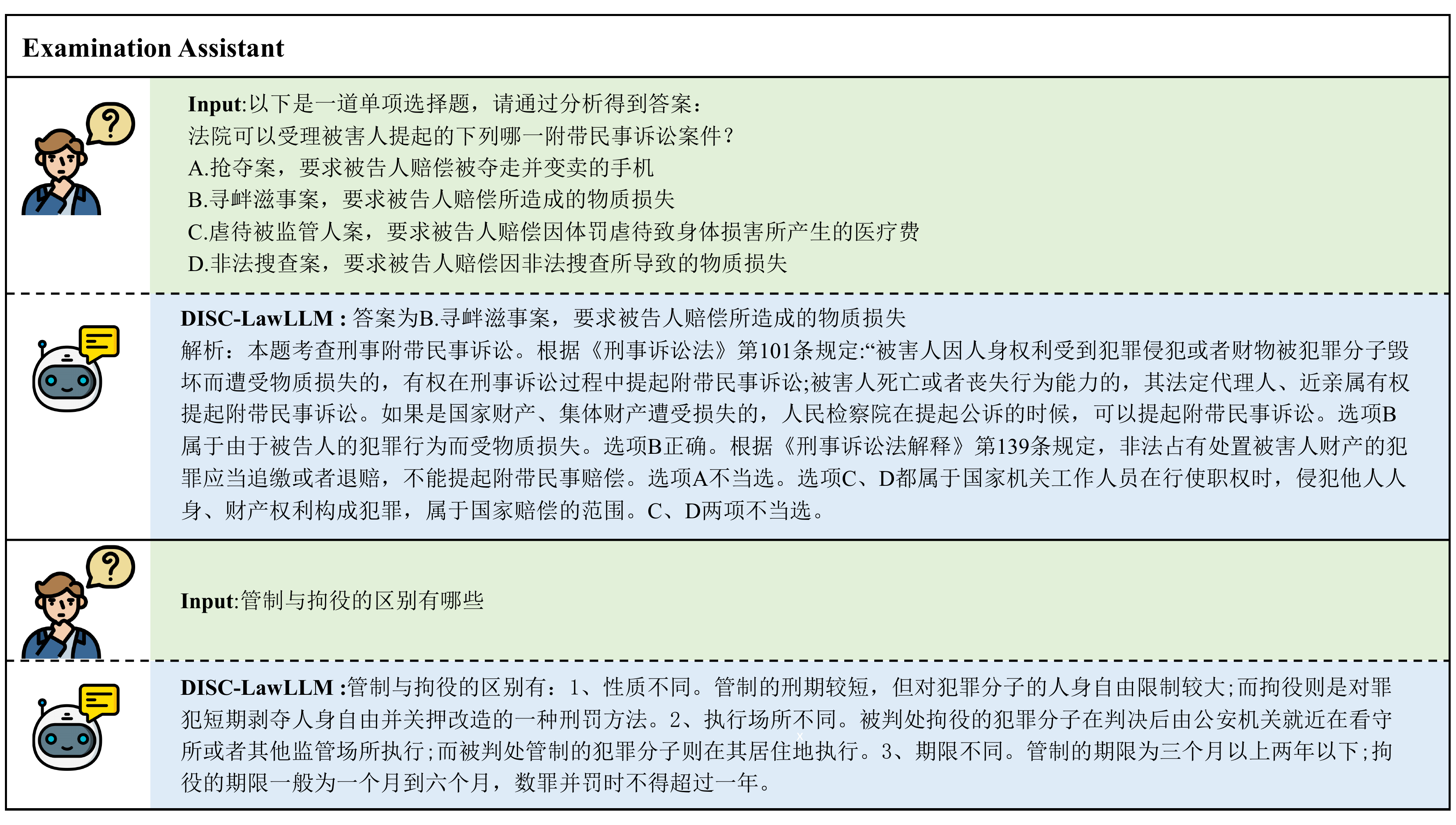}
    \caption{Case of our DISC-LawLLM as examination assistant}
    \label{fig:case1}
\end{figure*}

\begin{figure*}[!ht]
    \centering
\includegraphics[width=\linewidth]{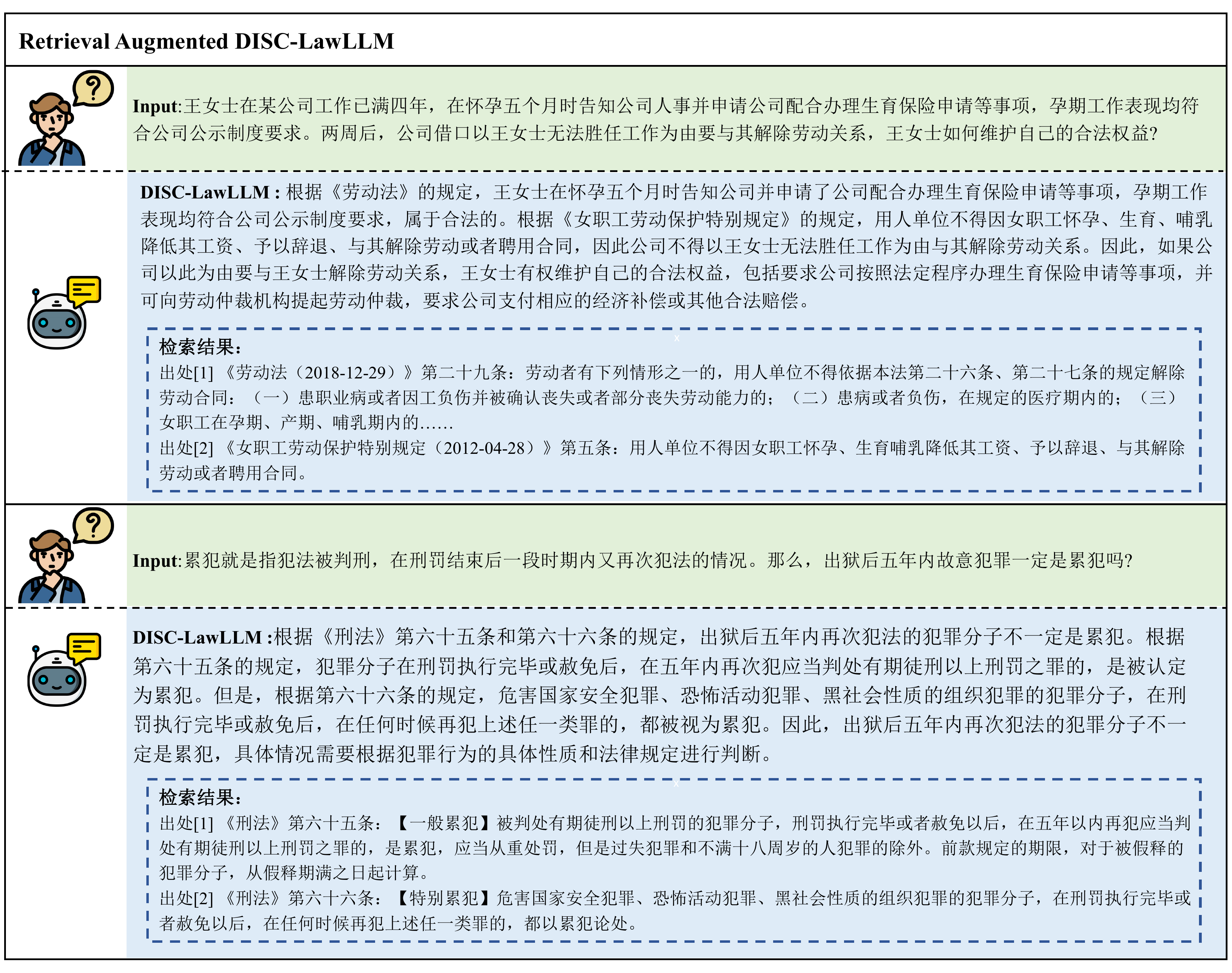}
    \caption{Cases of our Retrieval Augmented DISC-LawLLM.}
    \label{fig:case4}
\end{figure*}

\end{document}